# Risk Assessment Algorithms Based on Recursive Neural Networks

Alejandro Chinea *IEEE member*, Michel Parent

*Abstract—* The assessment of highly-risky situations at road intersections have been recently revealed as an important research topic within the context of the automotive industry. In this paper we shall introduce a novel approach to compute risk functions by using a combination of a highly non-linear processing model in conjunction with a powerful information encoding procedure. Specifically, the elements of information either static or dynamic that appear in a road intersection scene are encoded by using directed positional acyclic labeled graphs. The risk assessment problem is then reformulated in terms of an inductive learning task carried out by a recursive neural network. Recursive neural networks are connectionist models capable of solving supervised and non-supervised learning problems represented by directed ordered acyclic graphs. The potential of this novel approach is demonstrated through well predefined scenarios. The major difference of our approach compared to others is expressed by the fact of learning the structure of the risk. Furthermore, the combination of a rich information encoding procedure with a generalized model of dynamical recurrent networks permit us, as we shall demonstrate, a sophisticated processing of information that we believe as being a first step for building future advanced intersection safety systems.

## I. INTRODUCTION

IN the last few years a substantial effort of the European Union has been devoted to the organization of resources and the development of a sustainable approach to the prevention of traffic accidents. Moreover, it has been shown [1] that traffic accidents have become one of the major causes of death and injury not only in Europe but also globally. In addition, road intersections have been identified as one of the principal scenarios where traffic accidents take place. Several initiatives resulting from a joint effort between the European Union and the automotive industry have recently emerged [2] [3]. In this line, the aim of INTERSAFE [4] [5], a subproject of the IP PReVENT [6] project is to improve safety and to reduce fatal collisions at intersections by providing appropriate assistance to the driver. The driver should be warned against potential right of way violations caused by himself or other road-users, including failing to stop at red lights and stop signs, or turning across the path of other traffic. This objective is achieved by the development of risk assessment and data fusion algorithms that use as inputs the information collected from a combination of sensors for detection of vehicles and any other object present in the scene together with sensors for localization of the host vehicle, namely the test vehicle equipped with the INTERSAFE technology.

Highly-risky situations at road intersections are a combination of multiple factors that are very difficult to evaluate. The time-varying and highly non-linear nature of the problem makes extremely difficult to properly model the embeddings of information through time. In this paper, we propose a different point of view to the problem of computing risk functions. Specifically, we interpret the risk as a geometrical object embedded into a space-time manifold. The geometry of the risk is determined by taking into account the combined actions and roles of the objects present at road intersections (i.e. vehicles, pedestrians, traffic lights, trees, buildings and so forth). To this end, a novel information encoding procedure was designed to extract the intrinsic relations between such objects. Specifically, road intersection scenes are encoded by using directed ordered positional acyclic labeled graphs. The nodes of the graphs are labeled in the sense that they contain sets of domain variables represented by an array of numerical and categorical features.

This representation permits to combine disparate source of information (i.e. time and a priori knowledge about the intersection) in a compact structure that allows expressing complex relations between information units that can vary in size, as opposed to feature-based approaches. Once the information has been represented in terms of labeled graphs, the problem of risk assessment is expressed in terms of an inductive learning task performed by a recursive neural network. Recursive neural networks [7][8][9] are highly non-linear adaptive models that are able to learn structured data (i.e. trees or graphs patterns) even if such data is noisy or incomplete. The main motivation of using neural networks is reflected by the fact of being models that are able to recognize patterns we cannot even define. This property is called recognition without definition and it enables intelligent systems (i.e. like humans) to generalize. Namely, the ability to recognize multitudes of unforeseen complex patterns without any ability to define them. In our application, the recursive neural network is trained using a conveniently generated pattern set where each pattern corresponds to a road intersection scene encoded by means of directed ordered acyclic labeled graphs.

The rest of this paper is organized as follows: In the next

Manuscript received January 31, 2007. INTERSAFE is a subproject of the IP PReVENT. This work is part of the 6[th] Framework Program funded by the European Commission. The partners of INTERSAFE thank the European Commision for supporting the work of this project.

The authors are with INRIA, BP 105, Le Chesnay Cedex, France, Tel. + 33-1-39-63-55-47, Fax: + 33-1-39-63-51-14, e-mail: Alejandro.Chinea_Manrique_De_Lara@inria.fr, Michel.Parent@inria.fr.

section, the problem we address is presented within the context of the INTERSAFE European research project. In sections II and III, a formal description of the problem and the proposed approach is sketched. The experimental results are covered in section IV. Finally, section V provides a summary of the present study and some concluding remarks.

## II. PROBLEM STATEMENT

The European research project INTERSAFE is aimed to the development of active safety systems to prevent accidents at road intersections. In order to identify the most relevant scenarios of traffic accidents at intersections, a detailed accident analysis was performed by using statistics of France, Germany and Great Britain. It was found that in France and Germany accidents at intersections plays a central role in traffic accident databases. In addition, three accident scenarios were identified. Namely, accidents caused by collisions with oncoming traffic while turning left (first scenario). Accidents caused by collisions with crossing traffic while turning into straight crossing an intersection (second scenario) and finally traffic light controlled intersection accidents. In particular, red light violation leads often to serious crashes not depending on the type of accident. As an important result, it was shown that most of the accident situations are caused by human errors like misinterpretation or loss of attention.

From the results obtained from traffic accident analysis, it is important to note that a detailed knowledge of the traffic situation is a precondition for an optimal driver-supporting active safety system. To this end, the INTERSAFE test vehicle (hereafter the host vehicle) is equipped with two laser-scanners, one video camera and additional communication systems. The video camera is used to process data about lane markings at the intersection. Furthermore, the laser-scanners collects data of natural landmarks as well as data about other objects and road users. The fused data of the laser-scanner and the video camera together with a detailed map of the intersection are used for localization of the host vehicle within the intersection. Based on this information the advanced safety functions must be designed to correctly interpret the situation to avoid a possible collision. In addition, the driver behaviour (eg. human reaction time) must also be considered when developing the risk assessment algorithm. The ultimate goal is to provide appropriate warnings to the driver to prevent red light or stop sign violations, warnings when turning to prevent collisions with objects in the driving path or from the sides and finally warnings regarding time and distance to the traffic lights. The warnings are presented to the driver by a simple human machine interface (HMI) acoustic and optic based.

However, risk assessment algorithms are not sufficiently known and are still under an early stage of research [10][11]. This fact can be explained in terms of the inherent complexity of the problem. Firstly, the risk assessment algorithm must be able to operate with uncertainty, that is to say, expected imperfections or data errors coming from the measurement system. Secondly, the time-varying nature of the problem makes extremely difficult to properly capture and model the embeddings of information through time. Furthermore, potential risk situations are a combination of multiple factors which are not completely known and are also very difficult to evaluate. In the following subsections the details of our approach are presented, together with some experimental results.

## III. PROPOSED SOLUTION

A whole variety of scientific and technical fields are characterized by the fact that patterns of information appear in terms of structured objects. In general, nature provides us with many examples of structured objects, that is to say, objects that can be composed of smaller objects which may be structured too. Possible examples are chemical structures, mathematical expressions or software source code. Moreover, road intersections can be viewed as structured domains, where the structures can be either static or dynamic. Bearing in mind the time-variant nature of the risk and its mutual relation with the interactions and roles of the objects present at road intersections (i.e. vehicles, pedestrians, traffic lights, trees, buildings, etc), we are interested in generating an information encoding procedure as rich as possible to capture the intrinsic relationships between such entities.

Labeled graphs are graphical representations of information that are richer when compared to feature based representations as they permit to express more complex relationships between variables and/or concepts. The nodes of the graph are labeled in the sense that they contain sets of domain variables. Let us denote the set of domain variables associated to any node of a labeled graph as the label space. The variables contained within the label space can be either categorical or numerical. In addition, the presence of an edge between two nodes in a labeled graph indicates that the variables contained in those nodes are related in some way. Therefore, a natural way of encoding the information in a road intersection is by using directed positional acyclic graphs (hereafter DPAGs). More specifically, we encode the domain variables together with the static aspects (i.e. a priori knowledge about the intersection) of road intersections within the label space of the graph and the dynamic aspects of the scene throughout the structure of the graph.

However, the encoding procedure must be designed in conjunction with a computational procedure for the processing of structured information. It is important to note that we are interested in computing a risk value from the available information that the measurement system manages at any given instant time. Recursive neural networks are statistical pattern recognition systems which are able to learn information represented in terms of directed ordered acyclic labeled graphs. They are an extension of recurrent neural networks which are suited for performing supervised and non-supervised learning on data structures with some recent applications [12][13][14]. In the next subsections, the details of our approach are presented.

*A. Information Encoding Procedure*

Let us introduce some definitions and notations. A graph is said to be ordered if a total order $\prec$ is defined on the edges leaving from each vertex. The skeleton of a labeled graph is the data structure obtained by ignoring all the labels, but retaining the topology of the graph. A generic class of graphs skeletons having maximum in-degree i and maximum out-degree o is denoted by the symbol $\#^{(i,o)}$. Once we have specified a label space $Y$ and a skeleton class $\#^{(i,o)}$, we may define the space of data structures with labels in $Y$ and topology in #. Such a space is denoted by $Y^{\#}$.

A super-class of DOAGs is the class of directed positional acyclic graphs (hereafter DPAGs) in which it is assumed that for each vertex v, a bijection $P: E \rightarrow N$ is defined on the edges leaving from v. DPAGs are a super-class of DOAGs in the sense that every DOAG is also a DPAG, but not viceversa. Bounded DPAGs, or k-DPAGs, are defined by restricting the range of P to the finite set $[1,k] \in N$, for some fixed k. The term positional means that each child has its own position within the set of children of a node. Given a DOAG D and $v \in vert(D)$, we denote by *ch[v]* the set of children of v and by *pa[v]* the set of parents of *v*. The in-degree of *v* is the cardinality of the set *pa[v]* and the out-degree of *v* is the cardinality of the set *ch[v]*.

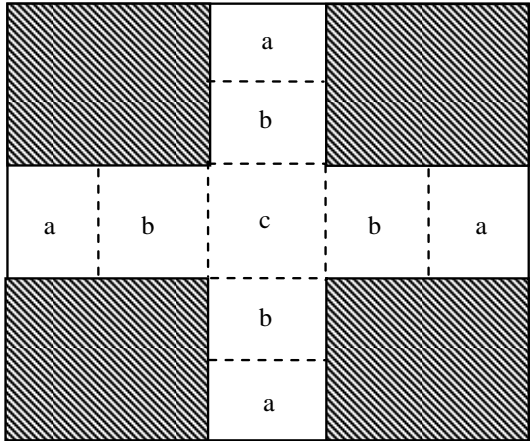

Fig. 1 Generic intersection subdivided into symmetrically equivalent zones.

The representation and encoding of information procedure can be stated as follows. Firstly, we perform a division of the intersection in zones, as suggested in [15] but with some substantial differences. We assign to each zone a set of predefined colors from a 6-tuple (yellow, orange, red, cyan, blue and white) that indicate the state of any dynamic object appearing in a scene. Secondly, the state of a dynamic object is defined in terms of its position and direction vector. In addition, directions are referred with respect to the center of the intersection. More specifically, regarding the coloring scheme, we consider the center of the intersection as the reference point to check the directions of dynamic objects. Furthermore, we just consider two generic directions. Namely, between two consecutive sampling points, dynamic objects can either increase either decrease their distance with respect to the center of the intersection. However, if the distance remains equal to zero, (i.e., The vehicle is stopped because of a traffic light) we just consider the previous state to associate a color to the current state. The reason to choose this coloring scheme is based on accident database statistics. Most of the accidents occur within the intersection or near-bound areas to the center of the intersection. Therefore, the encoding scheme cannot ignore this fact: Hot colors (i.e. yellow, orange, red) are assigned to states representing an increased probability of potential danger while cold colors (i.e. cyan, blue) represent just the opposite. Furthermore, colors representing similar concepts are closer from a hamming distance point of view (see table I).

The zone denoted with letter "a" of the intersection is denoted as *entering into the intersection* zone with yellow as associated color or *leaving the intersection* zone with blue as associated color depending if the distance of the dynamic object with respect to the center of the intersection is decreasing or increasing with time respectively. Similarly, the zone II is denoted as *approaching to the danger zo*ne with orange as associated color or *leaving the danger* zone with associated color cyan depending on the direction of movement of the dynamic object. Finally, the shaded area of the figure represents the concept of *out of the driving area*, to denote the portion of the intersection that is out of the road. This area contains for instance, pedestrians, bicycle paths, buildings, trees and so on.

TABLE I
DYNAMIC OBJECTS STATE SPACE REPRESENTATION

| Intersection Zone | Zone code | Associated Color | Encoding |
|---|---|---|---|
| Danger Zone | c | Red | 0 1 0 |
| Entering into intersection | a | Yellow | 0 0 1 |
| Approaching to the Danger zone | b | Orange | 0 1 1 |
| Leaving the danger zone | b | Cyan | 1 1 0 |
| Leaving the intersection | a | Blue | 1 1 1 |
| Out of the driving area | Shaded regions | White | 1 0 0 |

On the other hand, the label space was defined as the numerical and categorical information that we attach to the nodes as an array of features. We decompose this vector in three parts: the kinematic space, the state space of the dynamic object and finally the knowledge space. The kinematic space is composed by the set of variables used to characterize the movement of a dynamic object. In particular, we have used as domain variables: position, speed and direction. It is important to remember that at each sampling point the host vehicle recovers a perception of the environment expressed in terms of a set of measurement variables. The state space of the dynamic object is the set of states used to represent the behavior of dynamic objects (vehicles, pedestrians, bicycles, etc). In addition, the state of dynamic objects is represented by a 6-tuple of colors (see table I). Colors representing similar concepts are closer from a hamming distance point of view. Finally, the knowledge space represents the restrictions or a priori knowledge that can be associated to each graph node. In our experiments we just consider three knowledge dimensions: Speed limit violations, driving way violations and potential traffic light violations. These dimensions are represented by means of three binary variables (see table II).

TABLE II
ENCODING OF THE A PRIORI KNOWLEDGE IN THE LABEL SPACE

| Knowledge Space | Encoding |
|---|---|
| Driving Way Violation | 1 0 0 |
| Traffic Light Potential Violation | 0 1 0 |
| Speed Limit Violation | 0 0 1 |

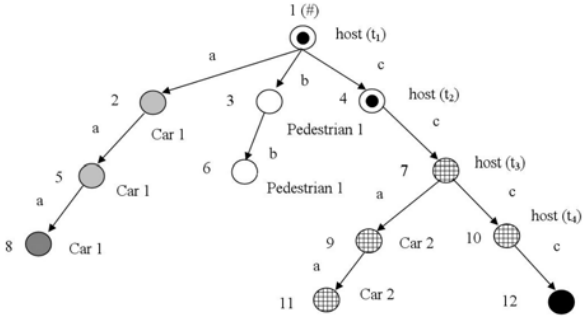

Fig. 2. An intersection scene encoded as a directed positional acyclic graph. Yellow state is represented by white circles with a smaller black circle inside. The Orange state is represented by checkered circles. Red state is represented by a black circle. Finally, cyan and blue states are represented by gray and dark gray circles respectively.

To give an example of the construction process, consider the graph depicted in Figure 2. The root node of the graph (denoted by the symbol #) will always correspond to the host vehicle. In addition, the maximum out-degree of the graphs is determined by the sensor characteristics, although from a practical point of view we can fix in advance this quantity to reduce the complexity of the graphs.

The position of the child indicates the type of object detected by the sensors of the host vehicle. In this line, a convention is defined to represent the host vehicle. Specifically, in our experiments the right-most child of any graph node is always assigned to represent the host vehicle (see figure 2). In this example we assume that the maximum out-degree of the graphs is three. Namely each node of the graph can have a maximum number of three children where each child corresponds to a dynamic object detected by the host vehicle with its sensors. The position of the child indicates the type of object detected by the host vehicle. In addition, the order in the graph is determined by the cardinality of the number attached to each node, which is related to the temporal sequence of events. Attached to each node there is also an informative label indicating the type of dynamic object (ie., vehicles, pedestrians, bicycles etc). The letters a,b,c denote the positions of any child with respect to its parents.

It is easy to deduce that the above-mentioned encoding procedure is invariant with respect to rotations and/or translations and naturally incorporate both symbolic and numerical information. Furthermore, the proposed encoding procedure does not depend on specific traffic regulation laws. For instance, driving ways may differ between the countries.

B. Recursive Neural Networks

Recursive Neural Networks are computational models capable of performing mappings from a set of labeled graphs to a set of real vectors. Specifically, they are based on the following recursive state space representation:

$$a(v) = f(a(ch[v]), I(v), v)$$
$$y(v) = g(a(v), v)$$

Here $v$ is a generic node identifier. The state vector $a(v)$ associated with node $v$ is a function of the ordered state vector associated with its children $a(ch[v])$ and the input vector $I(v)$ that encodes the set of attributes attached to node $v$ as a label.

The basic idea is to use recursive neurons to encode the structures. Namely, to perform the state transition function $f$ and the representations obtained are then used to approximate the output function $g$ by a standard feed-forward network, like a multilayer perceptron. The standard algorithm for training recursive neural networks is the Back-Propagation Through the Structure algorithm [16] (hereafter BPTS). However, BPTS suffers the problems of slow convergence associated to any gradient descent procedure. In this section we propose a quasi-newton learning algorithm hereafter Quasi-Newton Through the Structure (QNTS). Let us start by defining an error measure E(.) that is defined in the parameter space $\Omega_f \times \Omega_g$ which are represented by the model parameters $w^f$, $w^g$ (parameters of the state transition function $f$ and the output function $g$ respectively). In addition, we assume a training set of data structures $D = \{(U,Y), U \in U^\#, Y \in Y^\#\}$. The error function $E: \Omega_f \times \Omega_g \rightarrow R$ formulated for the inductive learning task, accumulates the error contributions of each input-output pair in the pattern set of structures $D$. Borrowing the notation of

[17] we can represent the standard squared error function as follows:

$$E = \sum_{(U,Y) \in D} E((U,Y)) = \sum_{v \in U} \frac{1}{2} \|y(v) - t(v)\|^2 \quad (1)$$

In the above equation the function t(.) is introduced to access the target label for a given node of the input graph. In order to devise a quasi-newton algorithm we need to compute the first-order gradient information to compute a sequence of matrices representing approximations of the

```
QNTS
Initialize Model parameters: W = W^f ∪ W^g, G = G^f ∪ G^g
W_0 ← random
G_0 ← P-GRADIENTS(D)
t ← 0
repeat
  if (t = 0)
    ΔW_0 ← -G_0
    H_0^{-1} ← I (Inverse Hessian initial aproximation using the identity matrix)
  else
    G_t ← P-GRADIENTS(D) (W is implicitly used)
    ΔG ← G_t - G_{t-1}
    ΔW ← W_t - W_{t-1}
    H_t^{-1} ← H_{t-1}^{-1} + ΔWΔW^T/ΔG^TΔG - H_{t-1}^{-1}ΔG ΔG^T H_{t-1}^{-1}/ΔG^T H_{t-1}^{-1} ΔG + (ΔG^T H_{t-1}^{-1} ΔG) u u^T
    ΔW_t ← -H_t^{-1} G_t (Update search direction)
    G_t ← G_{t-1}
  endif
  η_{t-1} ← Line Search(W_{t-1})
  W_t ← W_{t-1} - η_{t-1} H_{t-1}^{-1} G_{t-1}
  t ← t + 1 (Network performance is checked at this point)
until error is acceptable or maximum number of epochs reached
```

$$u = \frac{\Delta W}{\Delta W^T \Delta W} - \frac{H^{-1} \Delta G}{\Delta G^T H^{-1} \Delta G}$$

Fig. 3. Pseudo-code of the Quasi-Newton Through the Structure Algorithm

inverse Hessian (second order information). In addition, to reduce the effective complexity of the state transition network soft weight sharing technique is used. Namely, model parameters are shared among the nodes of the graphs and the gradient of a structural pattern $U$ is computed by summing up the individual node contributions. Let us denote by $w^f(v)$ and $w^g(v)$ the parameter vectors at node $v \in U$. Thus, we can express the error gradient of one graph $U$ with respect to the parameters $w^f$ and $w^g$ for $v \in U$. The parameters $w^g(v)$ do only influence the error contribution $E(v)$ of the node $v$. We get:

$$\frac{\partial E}{\partial w^g} = \sum_{v \in U} \frac{\partial E(U)}{\partial w^g(v)} = \sum_i \frac{\partial E(v)}{\partial y_i(v)} \frac{\partial y_i(v)}{\partial w^g(v)} = J_w^g(a(v))\delta^g(v) \quad (2)$$

Where $y_i(v)$ is the i-th component of the model output computed for node $v$, and $J_w^g$ is the Jacobian matrix of the output function g with respect to the parameters $w^g$ which has to be evaluated at location $a(v)$ and $\delta^g$ is given by $\delta^g(v) = y(v) - t(v)$. The error contribution of a node $z$ is influenced by all parameters $w^f(v)$, $v \in U$ where there is a path in the input graph leading from $z$ to nodes $v$. We have:

$$\frac{\partial E(U)}{\partial w^f} = \sum_i \frac{\partial E(v)}{\partial a_i(v)} \frac{\partial a_i(v)}{\partial w^f(v)} = J_w^f(a(ch[v]), I(v))\delta^f(v) \quad (3)$$

Where $a_i(v)$ is the i-th component of the state computed for node $v$, is the Jacobian matrix of the state transition function $f$ with respect to the parameters $w^f$ which is evaluated at the states $a(ch[v])$ known for the children of node $v$ and its input label $I(v)$. The term $\delta_i^f(v)$ is determined as follows:

$$\delta_i^f(v) = \frac{\partial E(U)}{\partial a_i(v)} = \sum_j \frac{\partial E(v)}{\partial y_j(v)} \frac{\partial y_j(v)}{\partial a_i(v)} + \sum_{n \in pd[v]} \sum_k \frac{\partial E(U)}{\partial a_k(n)} \frac{\partial a_k(n)}{\partial a_i(v)} \quad (4)$$

The above equation measures the effect of the state $a(v)$ on the error $E(v)$. The first term expresses the direct influence on $E(v)$ while the second one collects the indirect effects that are propagated through the parents of node v. Equation 4 can be expressed in matrix form as follows:

$$\partial^f(v) = J_x^g(a(v))\partial^g(v) + \sum_{n \in pa[v]} J_{x(r)}^f(a(ch[n], I(n)))\partial^f(n)$$

Where $J_x^g$ denotes the Jacobian matrix of the output function g with respect to its input arguments $J_{x(r)}^f$ stands for the transposed Jacobian of the state transition function $f$ to respect to those input arguments that belong to the corresponding child, for instance $r = ord(v,z)$ if the node $v$ is the r-th child of node $z$. From the considerations stated above, an algorithmic specification can be deduced. The pseudo-code of the quasi-newton algorithm is shown in figure 3. The algorithm works in batch mode. Namely, the gradients associated to the whole training set are computed before making an update on the parameter space of the neural network. Functions, S-GRADIENTS and P-GRADIENTS are used to compute the gradients associated to the structural patterns. The pseudo-code of these functions is presented in the appendix section. The procedure for computing the successive approximations to the inverse of the Hessian is BFGS [18]. In addition, the line search procedure used is backtracking [19].

IV. EXPERIMENTAL RESULTS

This section describes the experimental settings used to test the ideas presented so far. The goal pursued is just to show the potential of the approach throughout standard scenarios constructed by using simulated data.

Firstly, in our experiments, we consider a standard cross-like intersection consisting in a square of 100m$^2$, two driving ways, each way having two lanes of 3 meters each. We assume a speed limit of 80km/h when approaching to the intersection. The system coordinates, is placed at the center of the intersection. For the sake of simplicity we assume that the simulated scenes only contain two dynamic objects (host and remote vehicle). Therefore, the structural patterns belong to the skeleton class #$^{(1,2)}$ and are to be mapped by the recursive neural network into a continuous variable in the interval [0,1] indicating the level of risk. Similarly, from the accident scenarios explained in section II, we just consider the left turn across path and the traffic light scenario. In addition, in all the simulations the system is sampled at 200ms.

Moreover, with the aim of generating a realistic set of situations, a collision detection module was developed. Specifically, it is based on a trajectory module, a stochastic sampler and a collision detector. Trajectories are generated by using the following procedure: Firstly, a set of check-points in the kinematic space of the vehicles are generated which roughly describe the trajectories to be followed by the dynamic objects. For example, if the host vehicle is supposed to follow the lane and cross the intersection, some logical check-points would be: center point of the lane before the intersection, center point of the lane after the intersection and some exit point. Moreover, there is an error or uncertainty associated to each dimension of the kinematic space. Afterwards, a set of trajectories, constructed by using Bézier polynoms, are generated by sampling the stochastic check-points. In addition, speed, position and direction are modeled by using a gamma, triangular and normal probability distribution functions respectively. From the sampled trajectories, structural patterns are constructed by following the temporal sequence of the trajectories and shifting in time pairs of trajectories corresponding to the host and the remote vehicle. Thus, a whole variety of situations are simulated. In addition, visibility restrictions (i.e. sensors area coverage) are taken into account when generating the structural patterns. Moreover, the maximum depth of the tree-like graphs is limited by a risk window of 1.5s that takes into account the human reaction time.

TABLE III
EXPERIMENTAL RESULTS

| Architecture | Epochs | Collision Patterns Generalization (%) | Overall Generalization (%) |
|---|---|---|---|
| 23x160x1 | 500 | 99.4 | 97.9 |
| 20x150x1 | 500 | 100 | 94.2 |

Following the above-mentioned procedure a pattern set of 4000 structural patterns was generated. In addition, as stated in section III, the risk assessment problem is formulated as an inductive learning task. Therefore, two components are needed. Firstly, a teacher function is needed to rate the structural patterns. Specifically, the teacher provides a desired response or target for every structural pattern received from the dynamical system (the road intersection). In addition, the teacher function is supposed to be noisy as the quality of its associations will depend on the amount of knowledge encoded by the function. At this point, it is important to note that the expected goal is to produce a model that minimize the structural risk, outperforming at the same time the built-in knowledge of the teacher function, something that is not possible, if we think in a real-world industrial application, with simulated data. In our simulations, the teacher function is a weighted function that is mainly composed of two terms. The first term checks violations of the knowledge space while the second detects possible collisions of the host vehicle with the remote vehicle. Therefore, given a structural pattern the teacher function computes a value belonging to the interval [0,1] indicating the level of risk associated to the scene.

Secondly, a learning machine or algorithm is needed to learn the input-output mappings. As a learning machine recursive neural networks are used and trained with the algorithm proposed in section III. More specifically, from the pattern set, 2000 patterns are used for training and the other 2000 for validation purposes where 1138 of them corresponding to collision situations (target value equal to 1). Table III depicts the simulation results, two architectures were tested, the first having 23 state neurons, 160 hidden neurons and 1 neuron for the output layer. The second with 20 state neurons, 150 hidden neurons and 1 output neuron respectively. The output of the network and the targets associated to each structural pattern were discretized to compute the generalization capability of the networks using a threshold value of 0.5. Specifically, network outputs and targets bigger or equal than 0.5 were assigned to 1 and 0 respectively to indicate risk or not risk at all. In both simulations the algorithm was run 500 epochs and the generalization capability was tested at this moment. The results show that the model was able to recognize completely new dangerous situations. However, it must be noted that these results are just an indication of the potential of the ideas sketched throughout this paper. A strict validation of the proposed approach it is only possible with real data.

## V. CONCLUSIONS AND FUTURE WORK

In this paper we have investigated the application of recursive neural networks to compute risk functions. We have presented a formal description of the problem in terms of its application context. Specifically, within the Intelligent Transportation Systems research field.

The main contribution of this paper is reflected by the fact of considering the risk as a geometrical object with a specific structure that can be learnt by an appropriate computational model. More specifically, we have shown that the geometry of the risk is determined by taking into account the combined actions and roles of the objects present at road intersections. To this end, a novel information encoding procedure based on directed ordered acyclic graphs was proposed to extract the intrinsic relations between such objects. The proposed encoding procedure permitted us to combine disparate source of information in a compact structure that allowed expressing complex relations between information units that can vary in size.

Afterwards, the risk assessment at road intersections was reformulated in terms of an inductive learning task. At this point, recursive neural networks were introduced as highly non-linear adaptive models that are able to learn structured data even if such data was noisy or incomplete. The main drawback of these models is a computationally expensive training phase together with a laborious pattern generation

```
P-GRADIENTS(D)
  G ← 0
  repeat
    randomly select a pattern (U,Y) in the training set D
    G ← G + S-GRADIENTS(U,Y)
  until all patterns in the training set D have been selected
  return G

S-GRADIENTS(U, Y)
  Choose a reverse topological ordering ≻ on U
  for each node u ∈ U obeying the ordering ≻ do
    u.a ← f(ch[u], I(u))
  endfor
  for each u ∈ U do
    u.δ^f ← 0
  G^f ← 0, G^g ← 0
  Choose a topological ordering ≺ on U
  for each node u ∈ U obeying the ordering ≺ do
    δ^g ← y(u.a) − t(u)
    G^g ← G^g + J_w^g(u.a) δ^g
    u.δ^f ← u.δ^f + J_w^f(u.a)δ^g
    G^f ← G^f + J_w^f(a(ch[u]), I(u))·u.δ^f
    for v ∈ ch[u] do
      s ← ord(v,ch[u])
      v.δ^f ← v.δ^f + J_{w(s)}^f(a(ch[u]),I(u))·u.δ^f
    endfor
  endfor
  return G = G^g ∪ G^f
```

Fig. 4. Computation of the first derivatives of the error with respect to model parameters.

procedure. However, the main motivation of using neural networks is reflected by the fact of being models with the so-called recognition without definition property that enables systems to generalize. Specifically, the ability to recognize multitudes of unforeseen complex patterns without any ability to define them. Although further research must be carried out, the preliminary results have shown that the resulting model is able to recognize new dangerous situations given that risk information is implicitly encoded within the structural patterns that were learnt by the neural network during the training phase.

Our current work is focused on implementing the proposed methodology in a vehicle equipped with the INTERSAFE technology to validate the ideas presented throughout this paper in a real-time scenario.

## APPENDIX

The function S-GRADIENTS depicted in figure 4 expects a pair *(U,Y)* of input and target graph and returns the first-order gradients with respect to the model parameters $w^g$ and $w^f$. It is important to note that a topological ordering is defined for the structures. In addition, the function P-GRADIENTS it is just a helper function that simply collects the gradients contributions associated to each structural pattern in the training set.